\documentclass[sigconf]{acmart}

\usepackage{booktabs} 
\usepackage{algorithm, algpseudocode}
\usepackage{graphicx}
\usepackage{capt-of}
\usepackage{booktabs}
\usepackage{varwidth}

\usepackage{bbm,dsfont}
\usepackage{multirow}
\usepackage{amsthm}
\setcopyright{rightsretained}
\usepackage{algorithm}
\usepackage{multirow}
\usepackage{mathrsfs}
\usepackage{ragged2e}

\theoremstyle{definition}

\newcommand{\nonl}{\renewcommand{\nl}{\let\nl\oldnl}}

\acmDOI{10.475/123_4}

\acmISBN{123-4567-24-567/08/06}

\acmConference[SIGKDD'18]{}{Aug 2018}{London, United Kindom}
\acmYear{2018}
\copyrightyear{2018}

\acmPrice{15.00}

\begin{document}
\title{Interpretable Patient Mortality Prediction with Multi-value Rule Sets}

\author{Tong Wang}
\affiliation{%
  \institution{Tippie College of Business \\University of Iowa}
}
\email{tongwang@uiowa.edu}

\author{Veerajalandhar Allareddy}
\affiliation{%
  \institution{Carver College of Medicine\\University of Iowa}
}
\email{veerajalandhar-allareddy@uiowa.edu}

\author{Sankeerth Rampa }
\affiliation{%
  \institution{School of Business\\Rhode Island College}
}
\email{srampa@ric.edu}

\author{Veerasathpurush Allareddy}
\affiliation{%
  \institution{College of Dentistry\\
University of Illinois at Chicago}
}
\email{sath@uic.edu}

\renewcommand{\shortauthors}{T. Wang et al.}

\begin{abstract}
We propose a Multi-vAlue Rule Set (MRS) model for predicting in-hospital patient mortality. Compared to rule sets built from single-valued rules, MRS adopts a more generalized form of association rules that allows multiple values in a condition. Rules of this form are more concise than classical single-valued rules in capturing and describing patterns in data. 
Our formulation a`lso pursues a higher efficiency of feature utilization, which reduces possible cost in data collection and storage. We propose a Bayesian framework for formulating a MRS model and propose an efficient inference method for learning a maximum \emph{a posteriori}, incorporating theoretically grounded bounds to iteratively reduce the search space and improve the search efficiency.  Experiments show that our model was able to achieve better performance than baseline method including the current system used by the hospital.
\end{abstract}

%
%
%
%
%

\maketitle

\section{Introduction}\label{sec:intro}
In-hospital mortality prediction is an important problem for hospitals. Accurate predictions can help hospitals better allocate medical resources and provide targeted care for higher risk patients. Inpatient classification systems use mortality scores to classify patients into subgroups. But the scores are usually computed from black-boxes so doctors do not know how the decisions are made, which makes it difficult to gain trust and causes disagreement when the scores are inconsistent with doctors' expert judgement. Besides being uninterpretable, the inpatient classification systems are far from being satisfying in predictive performance. Here, our goal is to build an interpretable and accurate  model to predict in-hospital mortality of patients.

Among different forms of interpretable models, we investigate rule-based classifiers that  generate decisions based on a set of rules following simple ``if-then'' logic. Each rule is a conjunction of conditions made of features.
Prior rule-based models in the literature are built single value rules \cite{lakkaraju2016interpretable,wang2017bayesian, malioutov2013exact}.  For example, [Gender = Female] \emph{AND} [Age $\geq$ 40], where a \emph{condition} (e.g., [Gender = Female]) is a pair of a feature (e.g., gender) and a single value (e.g., female). 
However,  single valued rules are inadequate in capturing more general trends in the underlying data, especially when working with features with a high cardinality. 
For example, to capture a set of patients who have had procedure tracheostomy, or hemodialysis, or chest drainage, and the transfer status is to  a different acute care hospital or not transferred, a model needs six rules, each being a combination of procedure the patient has had and the transfer status,  
yielding a complicated and repetitive model. 

Another important aspect that has been overlooked by previous rule-based models is the need to control the total number of unique features. The number of different entities humans need to comprehend is directly associated with how easy it is to understand the model, as confirmed by the conclusions of  Miller \citep{miller1956magical} relative to the magical number seven. 
  With fewer features involved, it also becomes easier for domain experts to gain clear insights into data. 
In practice, models using fewer features are easier to understand and bring down the overall cost in data collection, especially in medical applications where collecting features incurs monetary costs, such as getting an examination result in a hospital

To combine the factors considered above, we propose a novel rule-based classifier, \emph{Multi-value Rule Sets} (MRS), which consists of rules that allow multiple values in a condition and uses a small set of features in the model. 
This form of model has great advantage over models built from classical single-value rules for (i) a more concise presentation of information and 
(ii) using a smaller number of features in the model.

We develop a Bayesian framework for learning MRS for its appealing property that it provides a unified
framework to jointly optimize fitting to the data and model
complexity without directly ``hard'' controlling either. We propose a principled objective combining interpretability and predictive accuracy where 
we devise a prior model that promotes a \emph{small} set of \emph{short} rules using a \emph{few} features. 
We propose an efficient inference algorithm for learning a maximum \emph{a posteriori} model.

\section{Related Work}

 There has been a series of research on rule-based models for classification \cite{ lakkaraju2016interpretable, pmlr-v56-Tran16, malioutov2013exact, rijnbeek2010finding}. 
These models use single-value rules, leaving presentation redundancy in the model. Also, most of these methods first generate a set of rules and then chooses a set from them to form the final model. This in practice will encounter the bottleneck of mining long rules. Furthermore, few of previous works consider limiting the number of features, and it is often independent of rule mining when they do. There has not been any work that combines rule learning and feature assignment into a unified framework.

 Our work is broadly related to generalized association rules that consider disjunctive relationships in items. Among various work along this line, some consider disjunction in the rule level, using the disjunction connector instead of conjunction connector as used by classical rule form. For example, $a_1 \vee a_2 \vee \cdots \rightarrow Y$, where $a_1$ is an item (e.g., gender $=$ female). Representative works include \cite{hilali2013mining, hamrouni2010generalization,hamrouni2009sweeping}. This primitive form of rules was extended to consider disjunctions in the condition/literal level\cite{nanavati2001mining}, yielding multi-value rules of the form $(a_1 \vee a_2 \vee \cdots) \wedge (b_1 \vee b_2 \cdots) \rightarrow Y$. 
 Prior efforts have mainly focused on mining individual multi-value rules \cite{srikant1995mining, hilali2013mining} using heuristics such as interestingness. Some works built classifiers comprised of multi-value rules \cite{berthold2003mixed,bombardier2007contribution}. However, they still rely on greedy methods such as greedy induction to build a model and they do not consider model complexity or restrict the number of features. Here, we optimize a global objective that considers predictive accuracy, model size, the the total number of features. By tuning the parameters in the Bayesian framework, our model can strike a nice balance between the different aspects of the model to suit the need of users.

\section{Multi-value Rule Sets}\label{sec:prem}
We work with standard classification data set $S$ that consists of $N$ observations $\{\mathbf{x}_n, y_n\}_{n=1}^N$. Let $\mathbf{y} \in \{0,1\}^N$ represent the set of labels and $\mathbf{x}$ represent the set of covariates $\mathbf{x}_n$. Each observation has $J$ features, and we denote the $j^\text{th}$ feature of the $n^\text{th}$ observation as $x_{nj}$.  Let $\mathcal{V}_j$ represent a set of values the $j^\text{th}$ feature takes. This notation can easily adapt to continuous attributes by discretizing the values into intervals.

\vspace{-2mm}
\subsection{Multi-value Rules}\label{sec:mvr}
 We introduce the basic components in MRS. 
 We define an \emph{item} as a pair of a feature $j$ and a value $v$, where $j \in \{1,2,\cdots,J\}$ and $v \in \mathcal{V}_j$.
A \emph{condition} is a collection of items with the same feature $j$,  denoted as $c = (j, V)$, where $j\in \{1,2,\cdots,J\}$ and $V\subset \mathcal{V}_j$. $V$ is a set of values in the item.
A \emph{multi-vale rule} is a conjunction of conditions, denoted as $r = \{c_k\}_{k}$.
Values can be grouped  into a value set.  In this way, a multi-value rule can easily describe the example mentioned in Section~\ref{sec:intro}. We only need one rule instead of six, yielding a more concise presentation while preserving the same information. Similarly for continuous features, multiple smaller intervals can be selected and merged into a more compact presentation.

[procedure = tracheostomy or hemodialysis or chest
drainage] AND [transfer status = not transferred or to a different
acute care hospital]

Now we define a classifier built from multi-value rules. By an abuse of notation, we use $r(\cdot)$ to represent a Boolean function that indicates if an observation satisfies rule $r$:
$r(\cdot):\mathcal{X} \rightarrow \{0,1\}.$
Let $R$ denote a Multi-value Rule Set. 
We define a classifier $R(\cdot)$:
$\mathbf{x}$ is classified as positive if it obeys \emph{at least} one rule in $R$ and we say $\mathbf{x}$ is \emph{covered} by $r$.

\vspace{-2mm}
\subsection{MRS Formulation}\label{sec:model}
The proposed framework considers two aspects: 1) interpretability, characterized by a prior for MRS, which considers the complexity (number of rules and lengths of rules) and feature assignment. 2) predictive accuracy, represented by the conditional likelihood of data given a MRS model. 

The prior model determines the number of rules $M$, lengths of rules $\{L_m\}_m$ and feature assignment $\{z_{m}\}_{m}$, where $m$ is the rule index and $z_m$ is a vector. We propose a  two-step process for constructing a rule set, where the first step determines the complexity a MRS model and the second step assigns the features.

\emph{Creating empty ``boxes'':} First, we draw the number of rules $M$ from a Poisson distribution with he arrival rate $\lambda_M$  determined via a Gamma distribution with parameters $\alpha_M, \beta_M$. Second, we determine the number of items in each rule, denoted as $L_m$. $L_m \sim \text{Poisson}(\lambda_L)$. The arrival rate, $\lambda_L$, is governed by a Gamma distribution with parameters $\alpha_L, \beta_L$.  Since we favor simpler models for interpretability purposes, we set $\alpha_L< \beta_L$ and $\alpha_M<\beta_M$ to encourage a small set of short rules. 
This step determines overall complexity to the model by creating empty ``boxes''.

\emph{Filling ``boxes':} Rule $m$ is a collection of $L_m$ ``boxes'', each containing an item. Let $z_{mk}$ represent the feature assigned to the $k^\text{th}$ box in rule $m$, where $z_{mk}\in\{1,...,J\}$ and $z_m$ represent the set of feature assignments in rule $m$. We sample $z_m$ from a  multinomial distribution with  weights $p$ drawn from a Dirichlet distribution parameterized by $\theta$.  Let $l_{mj}$ denote the number of items with attribute $j$ in rule $m$, i.e., $l_{mj} = \sum_k \mathbbm{1}(z_{mk} = j)$ and $\sum_{j}l_{mj} = L_m$. It means $l_{mj}$ items share the same feature $j$ and therefore can be merged into a condition. We truncate the multinomial distribution to only allow $l_{mj}\leq |\mathcal{V}_j|$.    
Here's the prior model, with parameters $H_s = \{\alpha_M, \beta_M,$ $\alpha_L, \beta_L, \theta\}$.
\begin{align}
&	M \sim \text{Poisson}(\lambda_M)& \lambda_M \sim \text{Gamma}(\alpha_M,\beta_M)& \\
&L_m \sim \text{Poisson}(\lambda_L), \forall m  &  \lambda_L \sim \text{Gamma}(\alpha_L,\beta_L) &\\
&z_{m}\sim \text{Multinomial}(p),\forall m   &p\sim \text{Dirichlet}(\theta)&
\end{align}
\paragraph{Conditional Likelihood}
 We consider the predictive accuracy of a MRS by modeling the conditional likelihood of  $\mathbf{y}$ given $\mathbf{x}$ and  model $R$. Our prediction on the outcomes are based on the coverage of MRS. 
We assume  label $y_n$ is drawn from Bernoulli distributions. Specifically, when $R(\mathbf{x}_n)=1$, i.e., $\mathbf{x}_n$ satisfies the rule set, $y_n$ has probability $\rho_+$ to be positive, and when $R(\mathbf{x}_n)=0$, $y_n$ has probability $\rho_-$ to be negative. 
We assume $\rho_+, \rho_-$ are drawn from two Beta distributions  with  hyperparameters $(\alpha_+,\beta_+)$ and $(\alpha_-, \beta_-)$, respectively, which control the predictive power of the model.  The conditional likelihood is as below given parameters $H_c = \{\alpha_+,\beta_+,\alpha_-,\beta_-\}$:
$
 p(\mathbf{y}|\mathbf{x}, R;H_c) \propto B(\text{tp}+\alpha_+,\text{fp}+\beta_+)B(\text{tn}+\alpha_-,\text{fn}+\beta_-),
$
where tp, fp, tn and fn represent the number of true positives, false positives, true negatives and false negatives, respectively. $B(\cdot)$ is a Beta function and comes from integrating out $\rho_+, \rho_-$ in the conditional likelihood function. 

\vspace{-2mm}
\subsection{Inference Method}
%
We devise an efficient inference algorithm that adopts the algorithm structure from \cite{wang2017bayesian}. 
Given an objective function $p(R|S)$ over discrete search space of different rule sets and a temperature 
schedule function over time steps, $T^{[t]} = T_0^{1 - \frac{t}{N_\text{iter}}}$, a simulated
annealing \cite{kirkpatrick1983optimization} procedure is a discrete time, discrete state Markov
Chain where at step $t$, given the current state $R^{[t]}$, the
next state $R^{[t+1]}$ is chosen by first proposing a neighbor and accepting it with  probability $\exp\left(\frac{p(R^{[t+1]}|S) - p(R^{[t]}|R)}{T^{[t]}}\right)$. 

A ``next state'' is proposed by first selecting an action to alter the current MRS and then choose from the ``neighboring'' models generated by that action. To improve search efficiency, we sample from misclassified examples to determine an action that can improve the current state $R^{[t]}$. 
If the misclassified example is positive, it means the $R^{[t]}$ fails to ``cover'' it and therefore needs to increase the coverage by randomly choose from adding a value, removing a condition, or adding a rule to the rule.
The three actions increase the coverage of a rule set.
On the other hand, if the misclassified example is negative, it means $R^{[t]}$ covers more than it should and therefore needs to reduce the coverage by randomly choosing from adding a condition or removing a rule from $R^{[t]}$
The two actions reduce the coverage of a model.

Performing an action involves choosing a value, a condition, or a rule. To select one from them, we evaluate the posterior on every model.  Then a choice is made between exploration (choosing a random model) and exploitation (choosing the best model). This randomness helps to avoid local minima and helps the Markov Chain to converge to a global optimum.

\vspace{-2mm}

\section{In-Hospital Mortality Prediction}

\textbf{Dataset} We used a real-world dataset from Nationwide Inpatient Sample (NIS). The NIS  is the largest all-payer hospitalizations database in the United States. It is a component of the Healthcare Cost and Utilization Project family of databases and is sponsored by the Agency for Healthcare Research and Quality. 
For our study, we examined the NIS database for the years 2012 to 2014 to obtain a cohort of patients aged 18 years and above who underwent Hematopoietic Stem Cell Transplant procedures during the hospitalization period. The dataset we worked on is comprised of   $9,451$ patient records, each described by $368$ features related to medical history, insurance status, geographic region, etc. The medical history includes a list of procedures a patient has received, measure of comorbidity burden, chronic conditions, a set of diagnosis, and etc. Here, $y_i=1$ if a patient died in hospital. This dataset is imbalanced with 2.9\% of positives. 

In addition to these features, the dataset also contains \emph{APRDRG Risk Mortality} and \emph{APRDRG severity} from an inpatient classification system widely adopted by hospitals in the US, APRDRG.  It represents All Patient Refined Diagnostic Related Group.  The two APRDRG scores categorize the severity of illness and risk of mortality into four subclasses, minor, moderate, major, and extreme.  They are computed taking into account several variables such as principal diagnosis, age, interactions of multiple secondary diagnoses, and combinations of non-operating procedures with principal diagnosis.  They can be used for risk-adjustment of outcomes and hospital casemix. Hospitals rely on these APRDRG scores to predict mortality. Here, we aim to build a more accurate model and investigate the predictive performance of APRDRG scores. 

\textbf{Experiments Setup}  We partition the dataset into 80\% training set and 20\% testing set. We benchmark the performance of MRS against the following rule-based models for classification: decision trees (C4.5 \cite{quinlan2014c4} and C5.0 \cite{quinlan2004data}), Repeated Incremental Pruning to Produce Error Reduction (Ripper) \citep{cohen1995fast}, Scalable Bayesian Rule Lists (SBRL) \citep{ynormalize_addang2016scalable}, and Bayesian Rule Sets (BRS) \citep{wangbayesian}. Among those, Ripper is one of the earlier rule-based classifier that was designed
to bridge the gap between association rule mining and classification  and focused mostly on optimizing for predictive accuracy. 
On the other hand, BRS and SBRL, two recently proposed frameworks
aim to achieve simpler models alongside predictive accuracy. Both use single-valued rules.  

We used R packages \cite{hornik2007rweka} for baseline methods except for BRS which has the code publicly available\footnote{\url{https://github.com/wangtongada/BOA}}.
To tune the parameters, we used 5-fold cross validation on the training set.  For both tree models, we tuned the minimum number in a leaf via cross-validation. For C5.0, we activated feature selection in the model, to help the model improve the performance.   For BRS, we set the maximum length of rules to 3. There are parameters $\alpha_+,\beta_+,\alpha_-,\beta_-$ that govern the likelihood of the data. We set $\beta_+, \beta_-$ to 1 and vary $\alpha_+,\alpha_-$ from $\{100,1000,10000\}$. For SBRL, we set the maximum length of rules to 2 since the computer will have a memory overflow using longer rules (the computer used for this experiments has 3.5 GHz 6-Core Intel Xeon E5 processor and 64 GB 1866 MHz DDR3 RAM). There are hyperparameters $\lambda$ for the expected length of the rule list and $\eta$ for the expected cardinality of the rules in the optimal rule list.  We vary $\lambda$ from $\{5,10,15,20\}$ and $\eta$ from $\{1,2,3,4,5\}$.
We kept BRS and SBRL models that have a total number of rules to be less than 5 for them to be comparable in complexity with the MRS model.
 
 Besides these models, we also evaluated the performance of the existing APR-DRG system. We tried different cut-off point from minor, moderate, major, and extreme and report the one with the highest F1 score on the test set.

\begin{table}[h!]
\centering
\caption{Performance comparison of different methods.}
\label{tab:mortality}
\begin{tabular}{@{\hskip3pt}l@{\hskip3pt}@{\hskip3pt}c@{\hskip3pt}@{\hskip3pt}c@{\hskip3pt}@{\hskip3pt}c@{\hskip3pt}@{\hskip3pt}c@{\hskip3pt}@{\hskip3pt}}
\toprule
                  & F1   &  $n_\text{rule}$ & $n_\text{cond}$         &  $n_\text{feat}$   \\\hline
C45            &  .45    & 17  &   39   & 14   \\
C50            & .39     & 5  &  11   &   5 \\
CART  &  .52    &20   &   39  & 15   \\
Ripper            & .50   &  3  &  12 & 11  \\
SBRL              &  .46  &   4   &  4    &  3  \\
BRS               & .46     & 1     & 3     & 3     \\
\textbf{MRS}              & .54 &  1  &   2    &   2   \\
\hline
APRDRG Risk Mortality&     .50 &1&1&1\\

\bottomrule 
\end{tabular}
\end{table}

\begin{table*}[h!]\label{tab:exp}
\centering
\small
\caption{The output MRS model.}
\begin{tabular}{ll}
\toprule
\textbf{if} &APRDRG Risk of Mortality = Extreme \emph{AND} procedure category = 33 or 34 or 35 or 39 or 58 or 61 or 63 or 142 or 216 or 225, \\ 
\textbf{then} &the patient is predicted to die in hospital, \\
\textbf{else} &the patient is predicted to not die in hospital.                                                                                                                                                                                                                                                                                                                                                                                                                                                                                      \\ \bottomrule
\end{tabular}
\end{table*}
\textbf{Results} Since the data is highly imbalanced, we choose F1 score as a metric for predictive performance and report the F1 scores on the testing data for each model in Table 1.  To compare the model complexity, we also report the number of rules (or leaves in decision trees), the number of conditions, and the number of features used by each model. 
Table 1 shows that MRS achieved a higher F1 than all baseline methods, while using  fewest rules, conditions, and features.  Methods that are comparable in predictive performance (e.g., CART) use significantly larger models built from more conditions, rules, and features. Models that are comparable in complexity (BRS and SBRL) lose too much in predictive performance. APRDRG scores, which are used by hospitals in the real world, turns out to be very competitive, only losing  0.04 to the MRS model.

A main reason for MRS's better performance is that many important features in the dataset, such as the procedures the patient has received, comorbid condition, historical diagnoses, etc, have high cardinalities. BRS and SBRL use single value rules. Therefore many promising rules have a very small support for only using one value for each condition, thus being neglected by the model. Decision trees tend to generate a large and complicated tree since a feature may be split multiple times to get meaningful regions in the data space.  MRS, on the other hand, does not have these issues. It includes multiple values in a condition.
We show the MRS model in Table 2. If a patient satisfies any rule in the model, s/he is predicted to incur mortality.

This model consists of only one rule with two conditions. 
The feature ``procedure'' represents the procedures the patient has taken.  They take multiple values in the model. The meaning of the values is shown in Table 3.
If we converted this model to a single-valued rule set, we would obtain a model with ten rules (each taking a combination of APRDRG Risk of Mortality = Extreme  and one procedure category), it will produce a total of 20 conditions, which is a more repetitive model representation than the MRS model we show here. 
This model also illuminates why APRDRG scores alone are not accurate enough - information from other features are missing. Therefore when refined with other features (see rules in the second model), MRS was able to achieve better predictive performance.

\begin{table}[h!]\label{tab:procedures}
\centering
\small
\caption{Descriptions of features used in the MRS model.}
\begin{tabular}{ll}
\toprule
Category labels & Description \\
 \hline
33 &  Cancer of kidney and renal pelvis \\
34 & Cancer of other urinary organs \\
35 &Cancer of brain and nervous system \\
39 & Leukemias \\
58 & Other nutritional; endocrine; and metabolic disorders  \\
61 &  Sickle cell anemia \\
63 &  Diseases of white blood cells \\
142 & Appendicitis and other appendiceal conditions \\
216 & Nervous system congenital anomalies \\
225 & Joint disorders and dislocations; trauma-related                                                                                                                                                                                                                                                                                                                                                                                                                                                                                                    \\ \bottomrule
\end{tabular}
\end{table}

\section{Conclusions}
We built a Multi-value Rule Set (MRS) model to predict patient mortality. MRS has a a more concise and feature-efficient model form to classify and explain, each rule being a reason for the classification. Experiments on the real world dataset demonstrated that compared with classic and state-of-the-art rule-based models, MRS showed competitive predictive accuracy while achieving a significant reduction in complexity and the total number of features used, thus improving the interpretability. The output model also outperforms the APRDRG systems that are currently used in hospitals.

\noindent\textbf{Code:} See the code at \url{https://github.com/wangtongada/MARS}.

\bibliographystyle{abbrv}
\bibliography{nips_msr,jmlr_msr,rules_ICDM}

\begin{thebibliography}{10}

\bibitem{berthold2003mixed}
M.~R. Berthold.
\newblock Mixed fuzzy rule formation.
\newblock {\em International journal of approximate reasoning}, 32(2-3):67--84,
  2003.

\bibitem{bombardier2007contribution}
V.~Bombardier, C.~Mazaud, P.~Lhoste, and R.~Vogrig.
\newblock Contribution of fuzzy reasoning method to knowledge integration in a
  defect recognition system.
\newblock {\em Computers in industry}, 58(4):355--366, 2007.

\bibitem{cohen1995fast}
W.~W. Cohen.
\newblock Fast effective rule induction.
\newblock In {\em Proceedings of the twelfth international conference on
  machine learning}, pages 115--123, 1995.

\bibitem{hamrouni2009sweeping}
T.~Hamrouni, S.~B. Yahia, and E.~M. Nguifo.
\newblock Sweeping the disjunctive search space towards mining new exact
  concise representations of frequent itemsets.
\newblock {\em Data \& Knowledge Engineering}, 68(10):1091--1111, 2009.

\bibitem{hamrouni2010generalization}
T.~Hamrouni, S.~B. Yahia, and E.~M. Nguifo.
\newblock Generalization of association rules through disjunction.
\newblock {\em Annals of Mathematics and Artificial Intelligence},
  59(2):201--222, 2010.

\bibitem{hilali2013mining}
I.~Hilali, T.-Y. Jen, D.~Laurent, C.~Marinica, and S.~B. Yahia.
\newblock Mining interesting disjunctive association rules from unfrequent
  items.
\newblock In {\em International Workshop on Information Search, Integration,
  and Personalization}, pages 84--99. Springer, 2013.

\bibitem{hornik2007rweka}
K.~Hornik, A.~Zeileis, T.~Hothorn, and C.~Buchta.
\newblock Rweka: an r interface to weka.
\newblock {\em R package version}, pages 03--4, 2007.

\bibitem{kirkpatrick1983optimization}
S.~Kirkpatrick, C.~D. Gelatt, M.~P. Vecchi, et~al.
\newblock Optimization by simulated annealing.
\newblock {\em science}, 220(4598):671--680, 1983.

\bibitem{lakkaraju2016interpretable}
H.~Lakkaraju, S.~H. Bach, and J.~Leskovec.
\newblock Interpretable decision sets: A joint framework for description and
  prediction.
\newblock In {\em ACM SIGKDD}, pages 1675--1684. ACM, 2016.

\bibitem{malioutov2013exact}
D.~Malioutov and K.~Varshney.
\newblock Exact rule learning via boolean compressed sensing.
\newblock In {\em International Conference on Machine Learning}, pages
  765--773, 2013.

\bibitem{miller1956magical}
G.~A. Miller.
\newblock The magical number seven, plus or minus two: some limits on our
  capacity for processing information.
\newblock {\em Psychological review}, 63(2):81, 1956.

\bibitem{nanavati2001mining}
A.~A. Nanavati, K.~P. Chitrapura, S.~Joshi, and R.~Krishnapuram.
\newblock Mining generalised disjunctive association rules.
\newblock In {\em Proceedings of the tenth international conference on
  Information and knowledge management}, pages 482--489. ACM, 2001.

\bibitem{quinlan2014c4}
J.~R. Quinlan.
\newblock {\em C4. 5: programs for machine learning}.
\newblock Elsevier, 2014.

\bibitem{quinlan2004data}
R.~Quinlan.
\newblock Data mining tools see5 and c5. 0.
\newblock 2004.

\bibitem{rijnbeek2010finding}
P.~R. Rijnbeek and J.~A. Kors.
\newblock Finding a short and accurate decision rule in disjunctive normal form
  by exhaustive search.
\newblock {\em Machine learning}, 80(1), 2010.

\bibitem{srikant1995mining}
R.~Srikant and R.~Agrawal.
\newblock Mining generalized association rules.
\newblock 1995.

\bibitem{pmlr-v56-Tran16}
T.~Tran, W.~Luo, D.~Phung, J.~Morris, K.~Rickard, and S.~Venkatesh.
\newblock Preterm birth prediction: Stable selection of interpretable rules
  from high dimensional data.
\newblock In {\em Proceedings of the 1st Machine Learning for Healthcare
  Conference}, volume~56 of {\em Proceedings of Machine Learning Research},
  pages 164--177, Northeastern University, Boston, MA, USA, 18--19 Aug 2016.
  PMLR.

\bibitem{wang2017bayesian}
T.~Wang, C.~Rudin, F.~Doshi, Y.~Liu, E.~Klampfl, and P.~MacNeille.
\newblock A bayesian framework for learning rule sets for interpretable
  classification.
\newblock {\em Journal of Machine Learning Research}, 2017.

\bibitem{wangbayesian}
T.~Wang, C.~Rudin, F.~Velez-Doshi, Y.~Liu, E.~Klampfl, and P.~MacNeille.
\newblock Bayesian rule sets for interpretable classification.
\newblock {\em ICDM}, 2016.

\bibitem{ynormalize_addang2016scalable}
H.~Yang, C.~Rudin, and M.~Seltzer.
\newblock Scalable bayesian rule lists.
\newblock {\em ICML}, 2017.

\end{thebibliography}

\end{document}